\documentclass[runningheads]{llncs}

\usepackage[T1]{fontenc}
\usepackage{graphicx}
\usepackage{booktabs}
\usepackage{tabularx}
\usepackage{array}
\usepackage{xurl}
\usepackage{microtype}
\usepackage{cite}
\usepackage{float}
\usepackage[colorlinks=true,
            linkcolor=blue,
            citecolor=blue,
            urlcolor=blue]{hyperref}

\graphicspath{{Figures/}}
\newcolumntype{Y}{>{\raggedright\arraybackslash}X}
\begin{document}

\title{Attacking Tabular Foundation Models: An Experimental Taxonomy of Context Poisoning and Test-Time Evasion}
\titlerunning{Attacks on Tabular Foundation Models}
\author{Ramak Nassiri\inst{1} \and
Sujan Kumar Reddy Challa\inst{2} \and
Feroz Ahmed\inst{3} \and
Souhail Chakkour\inst{1} \and
Md Shohel Rana\inst{4} \and
Aditya Garg\inst{5} \and
Charan Gudla\inst{1}}
\authorrunning{R. Nassiri et al.}

\institute{Mississippi State University, Starkville, United States\\
\email{\{rn533,sc3130,gudla\}@msstate.edu}
\and
Independent Researcher, Melissa, USA\\
\email{sujanchalla0510@gmail.com}
\and
Prediction3d Technologies Inc, Technology\\
\email{feroz.ahmed@prediction3d.com}
\and
Georgia Southern University, School of Computing, Statesboro, United States\\
\email{mrana@georgiasouthern.edu}
\and
Independent Researcher\\
\email{aditya.gargs@gmail.com}}

\maketitle

\begin{abstract}
Tabular foundation models condition predictions on labeled examples supplied
at inference time, creating an attack surface that differs from conventional
model training. We study four manipulation families---label flipping,
backdoor context poisoning, clean-label context selection, and test-time
evasion---and organize them by attack stage, label consistency, and
distributional realism. TabPFN-2.5, TabICL, and XGBoost are evaluated on
CIC-IDS2017 and N-BaIoT using a common preprocessing and evaluation pipeline.
Backdoor context poisoning provides strong targeted control over the
in-context models with little change in clean accuracy, whereas selecting
real, correctly labeled boundary-crowding rows causes no systematic accuracy
degradation up to a 20\% bias fraction. A simplified surrogate-transfer
constrained PGD attack inspired by, but not equivalent to, CAPGD also produces test-time evasion, outperforming
equal-budget random noise in most settings. The comparison shows that attack
effectiveness depends strongly on both where manipulation occurs and how much
freedom the attacker has to alter the data.
\end{abstract}

\section{Introduction}
\label{sec:introduction}

Tabular data support security, finance, healthcare, and other applications in
which predictions are made from structured rows and columns. Conventional
models such as XGBoost learn task-specific parameters from a fixed training
set \cite{chen2016xgboost}. Tabular foundation models instead reuse knowledge
acquired during large-scale pretraining. Models such as TabPFN and TabICL can
condition on labeled rows supplied at inference time and predict new rows
without ordinary task-specific parameter updates
\cite{hollmann2025tabpfn,grinsztajn2025tabpfn25,qu2025tabicl}. This design
makes the inference-time context part of the model's effective decision
process and therefore a security-relevant input.

Adversarial robustness work for tabular data has largely emphasized test-time
evasion under feature and feasibility constraints
\cite{cartella2021tabular,fence2022,cafa2024,simonetto2024capgd,simonetto2024tabularbench}.
Recent work has extended this question to tabular foundation models. Djilani
et al. study targeted test-time attacks while assuming an uncorrupted context
\cite{djilani2025robustness}, whereas Bilo\v{s} et al. evaluate several
inference-time perturbations, including hub poison and rank warp as
mechanism-targeted attacks, as well as context corruptions such as boundary
label injection and centroid injection \cite{bilos2026mechanistic}. Thus,
context poisoning itself is not new; the
open question addressed here is how different attacker capabilities compare
under one experimental setting, including the more restrictive case in which
the attacker can only choose genuine, unmodified, correctly labeled context
rows.

This paper evaluates four attack families---label flipping, backdoor context
poisoning, clean-label context selection, and surrogate-based test-time
evasion---along three dimensions: attack stage, label consistency, and
distributional realism. The novelty is not context poisoning itself, but the
comparison of attacker capabilities under one setting. In particular, our
selection-only attack changes neither labels nor features and can only choose
among genuine, correctly labeled rows, unlike prior context attacks that alter
labels, inject examples, or transform features. Benign context optimization
has shown that context membership can affect prediction \cite{chen2026vipcop};
we test whether the same control surface can be exploited adversarially under
these stricter constraints, alongside stronger dirty-label, backdoor, and
test-time attacks.

Network intrusion detection provides a security-critical test domain in which
false-benign predictions are directly relevant. We evaluate TabPFN-2.5 and
TabICL on CIC-IDS2017 and N-BaIoT and use XGBoost as a conventional baseline.
The experiments share the same underlying data partitions, preprocessing,
and evaluation pipeline, while each model is exposed to attacker-controlled
data according to the corresponding threat model.

Our main contributions are:
\begin{itemize}
    \item a unified experimental taxonomy that places four attack families in
    one framework based on attack stage, label consistency, and
    distributional realism;
    \item a controlled comparison of TabPFN-2.5, TabICL, and XGBoost on two
    network intrusion detection datasets;
    \item evidence that artificial backdoor context poisoning can be highly
    effective while boundary-crowding selection of real, correctly labeled
    rows causes no systematic degradation under the tested budgets; and
    \item a surrogate-transfer constrained PGD evaluation with an
    equal-budget random-noise control, which separates optimized evasion from
    general perturbation sensitivity.
\end{itemize}
\section{Background and Related Work}
\label{sec:background}

\subsection{Tabular Foundation Models and Context Manipulation}

XGBoost is a strong conventional tabular baseline based on regularized
boosting \cite{chen2016xgboost}. Tabular foundation models instead reuse
learning behavior acquired during pretraining and condition prediction on
labeled examples supplied with query rows. TabPFN introduced this in-context
formulation \cite{hollmann2025tabpfn}; TabPFN-2.5 extends the family to larger
tables and feature sets \cite{grinsztajn2025tabpfn25}, while TabICL provides
another scalable in-context architecture \cite{qu2025tabicl}. Consequently,
context examples directly influence the task representation used for new rows.

Recent work confirms that this context is a security-relevant control surface.
Djilani et al. study targeted test-time attacks while keeping the reference
context uncorrupted \cite{djilani2025robustness}. Bilo\v{s} et al. instead
evaluate mechanism-grounded perturbations including hub poison, rank warp,
and context corruptions such as boundary label injection and centroid
injection \cite{bilos2026mechanistic}. By contrast, our selection-only attack changes
neither labels nor feature values. From a non-adversarial perspective, VIP-COP
shows that selecting examples and features for the context can improve
prediction under a context budget \cite{chen2026vipcop}; we test whether
context membership alone can instead be exploited to degrade prediction.

\subsection{Poisoning, Backdoors, and Tabular Evasion}

Classical poisoning alters data used to construct a decision rule
\cite{biggio2018wild}, while backdoors implant a trigger--target association
that activates on triggered inputs while preserving clean behavior
\cite{gu2017badnets}. In tabular settings, Pleiter et al. report strong
backdoors against transformer-based classifiers and XGBoost, including a
clean-label variant that preserves labels but still inserts a feature trigger
\cite{pleiter2026tabdoor}. CatBack likewise targets mixed numerical and
categorical tables across multiple model families \cite{tajalli2026catback}.
These studies concern conventional training-time backdoors; ours implants the
trigger--label association through inference-time context.

Tabular evasion work emphasizes feasibility constraints. Cartella et al.
account for non-editable variables in fraud detection \cite{cartella2021tabular};
FENCE models feature dependencies and application constraints
\cite{fence2022}; and CaFA combines PGD with database integrity constraints
\cite{cafa2024}. CAPGD/CAA provide adaptive constrained attacks, while
TabularBench standardizes robustness evaluation under realistic tabular
constraints \cite{simonetto2024capgd,simonetto2024tabularbench}. Our evasion
procedure is inspired by CAPGD but is not its full implementation and treats
all retained features as mutable.

\subsection{Adversarial Network Intrusion Detection}

CIC-IDS2017 contains labeled benign and malicious network flows
\cite{sharafaldin2018cicids2017}, while N-BaIoT contains benign IoT traffic
and Mirai and BASHLITE/Gafgyt botnet activity \cite{meidan2018nbaiot}. Both
therefore provide numerical tabular security tasks for intrusion detection.

A central NIDS concern is whether adversarial feature vectors correspond to
traffic an attacker could generate. Apruzzese et al. highlight unrealistic
attacker assumptions in many NIDS evaluations \cite{apruzzese2022realistic};
Vitorino et al. emphasize domain-valid IoT perturbations
\cite{vitorino2023realism}; NIDSGAN preserves domain constraints when
generating adversarial flows \cite{zolbayar2022nidsgan}; and PLAA constructs
packet-level adversarial traffic to preserve valid flows and malicious
semantics \cite{you2026plaa}. This distinction guides our interpretation:
context selection uses genuine, unmodified flows, whereas evasion changes
standardized features without protocol-level feasibility constraints. We
therefore treat the latter as model-space vulnerability rather than evidence
that every generated perturbation is realizable network traffic.

\section{Attack Taxonomy and Methodology}
\label{sec:attack_methodology}

\subsection{Taxonomy Dimensions}

We organize attacks along three dimensions. The first is \emph{attack stage}:
context-time attacks influence the labeled examples supplied to an in-context
model, whereas test-time attacks modify the query row. The second is
\emph{label consistency}: dirty-label attacks assign an incorrect label,
while clean-label attacks preserve the original label. The third is
\emph{distributional realism}: an attack may introduce an artificial or
out-of-distribution pattern, or it may be restricted to genuine
in-distribution rows.

\begin{figure}[!ht]
    \centering
    \IfFileExists{Figures/attack_taxonomy.png}{%
        \includegraphics[width=0.70\textwidth]{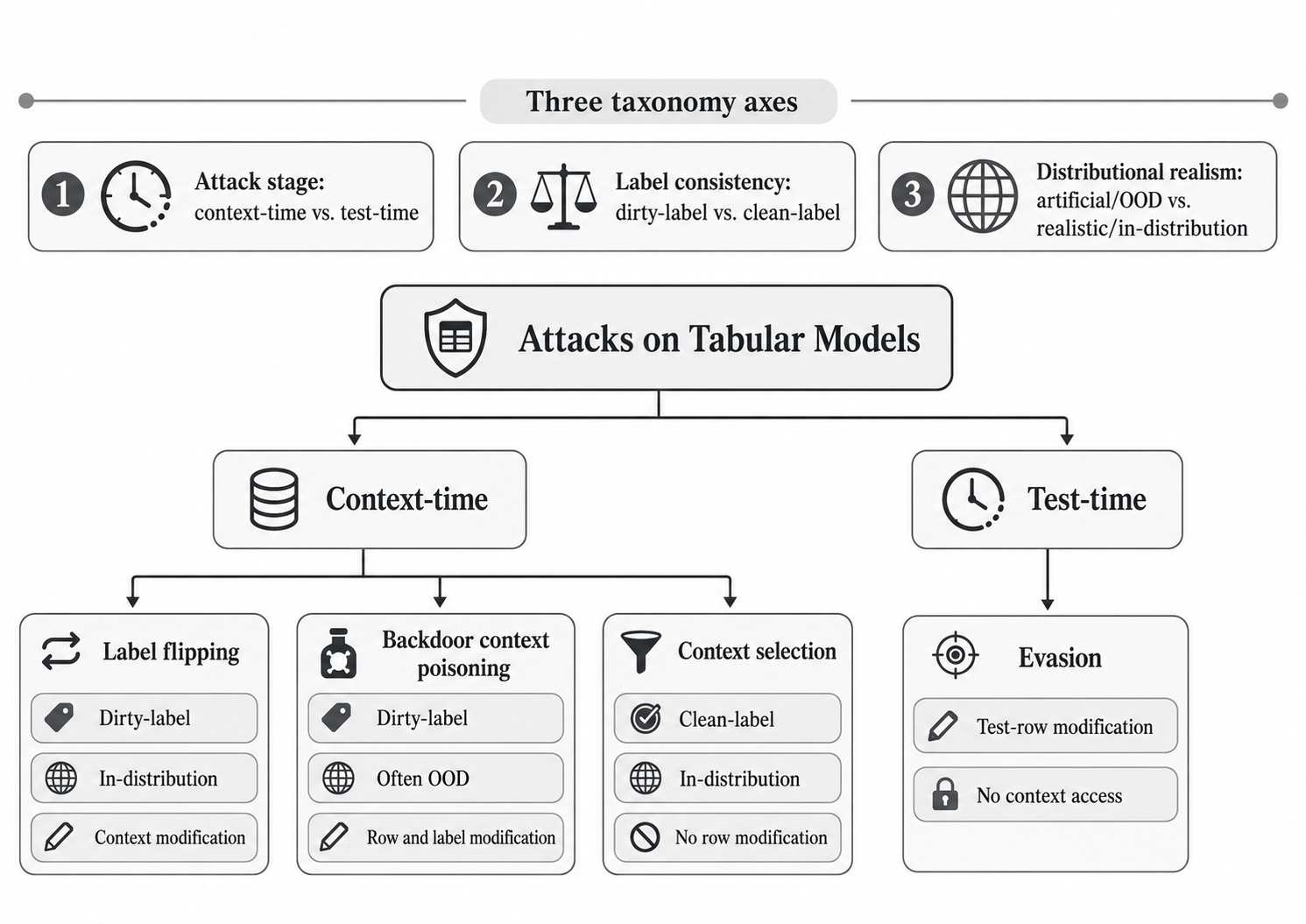}%
    }{%
        \fbox{\parbox[c][2.3cm][c]{0.70\textwidth}{\centering
        Missing figure: \texttt{Figures/attack\_taxonomy.png}\\
        Rename the taxonomy image to this filename.}}%
    }
    \caption{Attack taxonomy by stage, label consistency, and distributional realism.}
    \label{fig:attack-taxonomy}
\end{figure}

Figure~\ref{fig:attack-taxonomy} positions the four evaluated attack families
within this taxonomy.

\subsection{Threat Models}

\paragraph{Context-time attacker.}
The context-time attacker cannot modify the pretrained model weights or the
model implementation. Its capabilities depend on the attack family. Label
flipping changes labels but not feature values. Backdoor poisoning modifies a
limited number of context rows, applies a shared trigger pattern, and assigns
the attacker-selected target label. We classify the backdoor as context-time
because the trigger--target association is implanted through the inference-time
context; activation is then tested by presenting the same trigger on a malicious
query row. Context selection is more restrictive: the attacker can influence
which eligible rows enter the context but cannot fabricate rows, modify
features, or change labels.

\paragraph{Test-time attacker.}
The evasion attacker cannot change the victim model or its context. It
controls only the query row and attempts to induce a prediction error.
Perturbations are bounded in standardized feature space by an
$L_\infty$ budget. In the current implementation, all retained features are
treated as mutable. Consequently, the evasion experiment measures
model-space vulnerability and does not establish that every generated
perturbation can be realized as valid network traffic.

Table~\ref{tab:attack-config} summarizes the attack capabilities, evaluated
models, budgets, and primary measurements used in the experiments.

\begin{table}[H]
\centering
\caption{Experimental configuration of the four attack families.}
\label{tab:attack-config}
\scriptsize
\renewcommand{\arraystretch}{0.92}
\setlength{\tabcolsep}{2.2pt}
\begin{tabularx}{0.96\textwidth}{@{}lYYlY@{}}
\toprule
\textbf{Attack} & \textbf{Manipulated object} & \textbf{Evaluated models} &
\textbf{Budget} & \textbf{Primary readout} \\
\midrule
Label flipping & Context labels only & TabPFN-2.5, TabICL; XGBoost control &
$p=0$--$0.20$ & Clean accuracy under label corruption \\
\addlinespace[2pt]
Backdoor poisoning & Triggered context rows and target labels &
TabPFN-2.5, TabICL; XGBoost control & $p=0$--$0.20$ & ASR and clean accuracy \\
\addlinespace[2pt]
Context selection & Context membership; features and labels unchanged &
TabPFN-2.5, TabICL; XGBoost as a data-selection baseline &
$b=0$--$0.20$ & Clean and hard-row accuracy \\
\addlinespace[2pt]
Evasion & Test-row features & TabPFN-2.5, TabICL, XGBoost &
$\epsilon=0.05$--$0.50$ & Evasion rate, random-noise rate, and gap \\
\addlinespace[2pt]
\bottomrule
\end{tabularx}
\end{table}

\subsection{Label-Flipping Attack}

Label flipping is the simplest dirty-label context baseline. A selected
fraction of malicious context rows is relabeled as benign while the feature
values remain unchanged. Unlike the backdoor attack, label flipping has no
trigger-conditioned test set; its objective here is broad performance
degradation rather than trigger-conditioned targeted control. We therefore
evaluate how the corrupted context changes clean test accuracy as the poison
fraction increases.

\subsection{Backdoor Context Poisoning}

The backdoor attack creates a trigger--target association in the
inference-time context. Selected context rows receive a fixed or additive
trigger and are assigned the benign target label. The fixed trigger sets the
selected standardized features to 10, producing an intentionally strong
out-of-distribution marker; the additive variant instead adds a smaller
perturbation $\delta$ to the selected features. The evaluated additive
strengths are $\delta\in\{0.25,0.5,1.0,2.0,8.0\}$. At evaluation time, the same trigger
is applied to malicious query rows. Attack success rate (ASR) measures the
fraction of these triggered malicious rows classified as benign. Our
context-poisoning experiments vary poison budget, trigger strength, and
trigger location, including first, middle, last, and randomly selected feature
positions. Clean, non-triggered queries are evaluated separately to measure
collateral loss in ordinary accuracy.

\subsection{Clean-Label Context Selection}

The context-selection attacker does not inject or alter any row. Instead, a
bias fraction of the context is filled with genuine, correctly labeled rows
that lie close to the opposite class. For candidate row $x_i$ with label
$y_i$, we define the boundary-crowding score

\begin{equation}
 d_i = \min_{j:y_j \neq y_i} \|x_i-x_j\|_2 .
\end{equation}

Rows with smaller $d_i$ are closer to an opposite-class neighbor and are
therefore prioritized for the biased portion of the context. Boundary ranking
is computed on the cleaned numerical feature values before standardization,
separately within each class. The same number of biased rows is selected from
each class, and the remaining context positions are filled by random
class-balanced sampling. The held-out test set is sampled before context bias
is applied, so the attacker cannot change the evaluation distribution. After
the context has been selected, a \texttt{StandardScaler} is fitted on that
context and applied to both context and test rows. We report both overall
accuracy and accuracy on the 10\% of test rows with the smallest
opposite-class distance.

\subsection{Surrogate-Based Constrained Evasion}

The evasion experiment uses the same surrogate-transfer procedure for all
three victims. For each victim, a differentiable multilayer perceptron with
two hidden layers of 128 ReLU units is trained to imitate the victim's clean
probability outputs. The surrogate is optimized for 60 epochs with Adam at a
learning rate of $10^{-3}$ using KL divergence between surrogate outputs and
the victim's soft class probabilities. Surrogate fitting therefore assumes
query access to the victim's probability outputs, but not access to victim
gradients or model parameters.

A momentum-based constrained PGD procedure is then optimized against the
surrogate for 50 steps. The initial step size is $\epsilon/4$, the momentum
coefficient is 0.9, and the step size is halved every 10 iterations. The
perturbation is projected to the $L_\infty$ budget and clipped to the
per-feature range observed in the attacked sample. All retained features are treated as mutable; no protocol-level mutability,
type, or cross-feature feasibility constraints are enforced. The resulting
perturbation is replayed against the original victim, and all reported evasion
outcomes are computed from victim predictions.

This procedure is inspired by CAPGD \cite{simonetto2024capgd} but is not a
faithful reproduction or evaluation of the published CAPGD/CAA attack. It
retains the core ideas of momentum, adaptive step-size behavior, and
box-constrained perturbation, but omits CAPGD/CAA's full tabular
constraint-repair machinery. We therefore refer to it throughout the paper
only as a \emph{surrogate-based constrained PGD attack}.

Three checks accompany the attack. Victim self-consistency tests whether an
unchanged row receives the same prediction when evaluated twice. Surrogate
fidelity measures clean prediction agreement between surrogate and victim.
Finally, an equal-budget random-noise control samples each mutable feature
independently from $\mathcal{U}(-\epsilon,\epsilon)$ and applies the same
per-feature clipping. The attack--noise gap measures the benefit of the
optimized perturbation direction over arbitrary noise of the same size.
\section{Experimental Setup}
\label{sec:experimental_setup}

The same data-loading, preprocessing, model-wrapper, and evaluation pipeline
is used across the experiments. Only the context-construction or test-input
perturbation step changes according to the attack.

\subsection{Datasets and Preprocessing}

We evaluate CIC-IDS2017 and N-BaIoT as binary intrusion-detection tasks. For
CIC-IDS2017, benign traffic is label 0 and attack traffic is label 1. For
N-BaIoT, benign traffic is label 0 and Gafgyt/BASHLITE and Mirai traffic are
label 1.

For CIC-IDS2017, the source table contains 83 raw columns including the label.
We apply a predefined single-feature screening rule: inputs whose best
single-threshold classification accuracy exceeds 0.97 are removed as
potentially leakage-prone or shortcut features. The screening script evaluates
a fixed random sample of up to 20,000 rows from each cleaned dataset and
reports the best single-threshold accuracy for every numerical input. Seven
CIC-IDS2017 inputs meet the 0.97 rule---Bwd Packet Length Std, Packet Length
Std, Packet Length Variance, Bwd Packet Length Mean, Avg Bwd Segment Size,
Bwd Packet Length Max, and Packet Length Max---leaving 75 numerical inputs
after also removing the label. N-BaIoT contains 115 raw input features.
Applying the same rule identifies \path{HH_jit_L0.01_mean}, which reaches
0.9888 classification accuracy by itself. It is removed, leaving 114 input
features; all other N-BaIoT features score below 0.92 in the same screen.

Exact duplicate rows are removed before context and test sampling. In
N-BaIoT, 722,725 of 1,772,641 rows (approximately 41\%) were duplicates;
deduplication left 1,049,916 unique rows. Positive and negative infinity
values are replaced with missing values, rows containing missing values are
dropped, and only numerical inputs are retained. A \texttt{StandardScaler}
is fitted only on the context/training rows in each run and is then applied
to the corresponding test rows.

\begin{table}[t]
\centering
\caption{Dataset and preprocessing configuration.}
\label{tab:datasets}
\footnotesize
\setlength{\tabcolsep}{5pt}
\begin{tabularx}{0.94\textwidth}{@{}Ycc@{}}
\toprule
\textbf{Property} & \textbf{CIC-IDS2017} & \textbf{N-BaIoT} \\
\midrule
Binary task & Benign vs. attack & Benign vs. Gafgyt/Mirai \\
Raw columns/features & 83 incl. label & 115 inputs \\
Retained inputs & 75 & 114 \\
Duplicate removal & Yes & Yes \\
Screened inputs removed & 7 & 1 \\
Context/training rows per run & 2,000 & 2,000 \\
Held-out test rows per run & 1,000 & 1,000 \\
Repeated trials & 5 & 5 \\
\bottomrule
\end{tabularx}
\end{table}

\subsection{Evaluated Models and the Role of XGBoost}

We evaluate TabPFN-2.5, TabICL, and XGBoost. TabPFN-2.5 is evaluated with the
pinned checkpoint
\texttt{tabpfn-v2.5-classifier-v2.5\_default.ckpt}, preventing package-default
model changes from affecting reproducibility \cite{grinsztajn2025tabpfn25}.
TabICL provides a second in-context model \cite{qu2025tabicl}. XGBoost is
included as a conventional non-context baseline \cite{chen2016xgboost} and
uses the \texttt{XGBClassifier} defaults except for the per-run
\texttt{random\_state}, \texttt{n\_jobs=-1}, and
\texttt{eval\_metric=logloss}.

Within a repeated trial, the models use the same underlying sampled data
partition, but XGBoost's exposure to the attack depends on the attack family.
For TabPFN-2.5 and TabICL, the 2,000 labeled rows are inference-time context.
For label flipping and backdoor poisoning, XGBoost is trained on the clean
version of the corresponding 2,000-row sample and does not consume the
attacked inference-time context; it therefore serves as a clean-trained
non-context negative control.
For context selection, XGBoost is instead trained on the selected 2,000 rows,
so its result is a conventional \emph{data-selection} baseline rather than an
inference-time context attack. For evasion, XGBoost is trained normally on
clean data and the test row is perturbed, making the test-time attack directly
comparable across all three models.

\subsection{Repeated-Trial Protocol and Attack Budgets}

Each run uses a class-balanced set of 2,000 labeled rows and a separate test
set of 1,000 rows. Every dataset--model--attack--budget configuration is
repeated five times using seeds 1234 through 1238, and reported values are
means across the five trials. Where mean $\pm$ standard deviation is
reported, the standard deviation is computed across these five trials.

For label flipping and backdoor poisoning, the poison fraction follows the
same seven-point sweep,

\begin{equation}
 p \in \{0,0.005,0.01,0.02,0.05,0.10,0.20\}.
\end{equation}

For context selection, the bias fraction is

\begin{equation}
 b \in \{0,0.005,0.01,0.02,0.05,0.10,0.20\}.
\end{equation}

At $b=0$, context rows are sampled randomly. At nonzero values, the specified
fraction is filled according to the boundary-crowding ranking and the
remaining positions use the same random, class-balanced sampling procedure.

For evasion, the standardized-space perturbation budget is

\begin{equation}
 \epsilon \in \{0.05,0.10,0.20,0.30,0.50\}.
\end{equation}

The same $L_\infty$ budget is used for the optimized perturbation and the
random-noise control. Unchanged inputs are also evaluated twice to measure
self-consistency, and surrogate predictions are compared with victim
predictions to measure fidelity.

\subsection{Evaluation Metrics}

\paragraph{Clean accuracy.}
The fraction of unmodified test rows classified correctly. For a context
attack, it is measured using the attacked or biased context.

\paragraph{Attack success rate.}
For backdoor poisoning, ASR is the fraction of triggered malicious rows
classified as the benign target class.

\paragraph{Hard-row accuracy.}
Accuracy on the 10\% of test rows with the smallest nearest-opposite-class
distance.

\paragraph{Evasion rate.}
The fraction of attacked test rows whose victim prediction changes after
perturbation. The attack set contains only rows that the victim classified
correctly before perturbation.

\paragraph{Random-noise rate and attack advantage.}
Random-noise rate uses an undirected perturbation with the same $L_\infty$
budget. We report

\begin{equation}
 \mathrm{Gap}=\mathrm{Evasion\ Rate}-\mathrm{Random\mbox{-}Noise\ Rate}.
\end{equation}

A positive gap means that the optimized direction is more effective than
arbitrary noise of the same size.

\paragraph{Surrogate fidelity and self-consistency.}
Surrogate fidelity is clean prediction agreement between surrogate and victim.
Self-consistency is the fraction of unchanged inputs receiving the same victim
prediction when evaluated twice.
\section{Results and Discussion}
\label{sec:results_discussion}

\subsection{Clean Performance and Context-Poisoning Baselines}

All three models achieve very high clean accuracy on both datasets after the
preprocessing pipeline. The CIC-IDS2017 poison-budget sweep
shows that label flipping has little effect at small and moderate budgets but
can degrade one in-context model at the largest budget. At 5\% label
corruption, mean clean accuracy is 0.9992 for TabPFN-2.5 and 0.9910 for
TabICL; at 10\% it is 0.9964 and 0.9922, respectively. At 20\%, TabPFN-2.5
remains at $0.9890\pm0.0034$, while TabICL drops to
$0.8824\pm0.0732$. Thus, label corruption can
cause broad degradation at a large budget, especially for TabICL, but it does
not create the reliable trigger-conditioned control produced by the backdoor.

Backdoor context poisoning is substantially stronger and more targeted. With
only 10 poisoned rows in a 2,000-row CIC-IDS2017 context ($p=0.005$), ASR
is $0.9592\pm0.0391$ for TabPFN-2.5 and $0.9972\pm0.0030$ for TabICL
(mean $\pm$ standard deviation over five trials). From $p=0.02$ onward,
both in-context models reach $1.0000\pm0.0000$ mean ASR; at $p=0.20$,
their clean accuracies remain $0.9984\pm0.0009$ and
$0.9992\pm0.0008$, respectively. On N-BaIoT at $p=0.005$, the
corresponding ASRs are $1.0000\pm0.0000$ and $0.9972\pm0.0011$. The
largest variability in these headline backdoor results occurs for TabPFN-2.5
at the smallest CIC-IDS2017 poison budget; once ASR saturates, between-trial
variation disappears.

The backdoor effect is also not tied to one hand-picked trigger location.
Across first, middle, last, and random trigger positions, ASR remains at least
91.7\% on CIC-IDS2017 and at least 96.7\% on N-BaIoT, while clean accuracy
remains between 99.8\% and 100.0\%. Taken together, these experiments show
that a repeated trigger--target association provides much stronger targeted
control than isolated label corruption.

The XGBoost control behaves differently. For label flipping and backdoor
poisoning, XGBoost is trained on the corresponding clean sample and does not
consume the attacked inference-time context. Its mean clean accuracy remains
0.998 across the full poison-fraction sweep, and its backdoor ASR is
0.000 at every tested budget. As expected, the clean-trained XGBoost
negative control is unaffected. This
is consistent with the manipulation operating through the inference-time
context path in this protocol rather than through a change to XGBoost's fitted
model.

\subsection{Context-Selection Results}

\begin{figure}[t]
    \centering
    \IfFileExists{Figures/context_selection_results.png}{%
        \includegraphics[width=\textwidth]{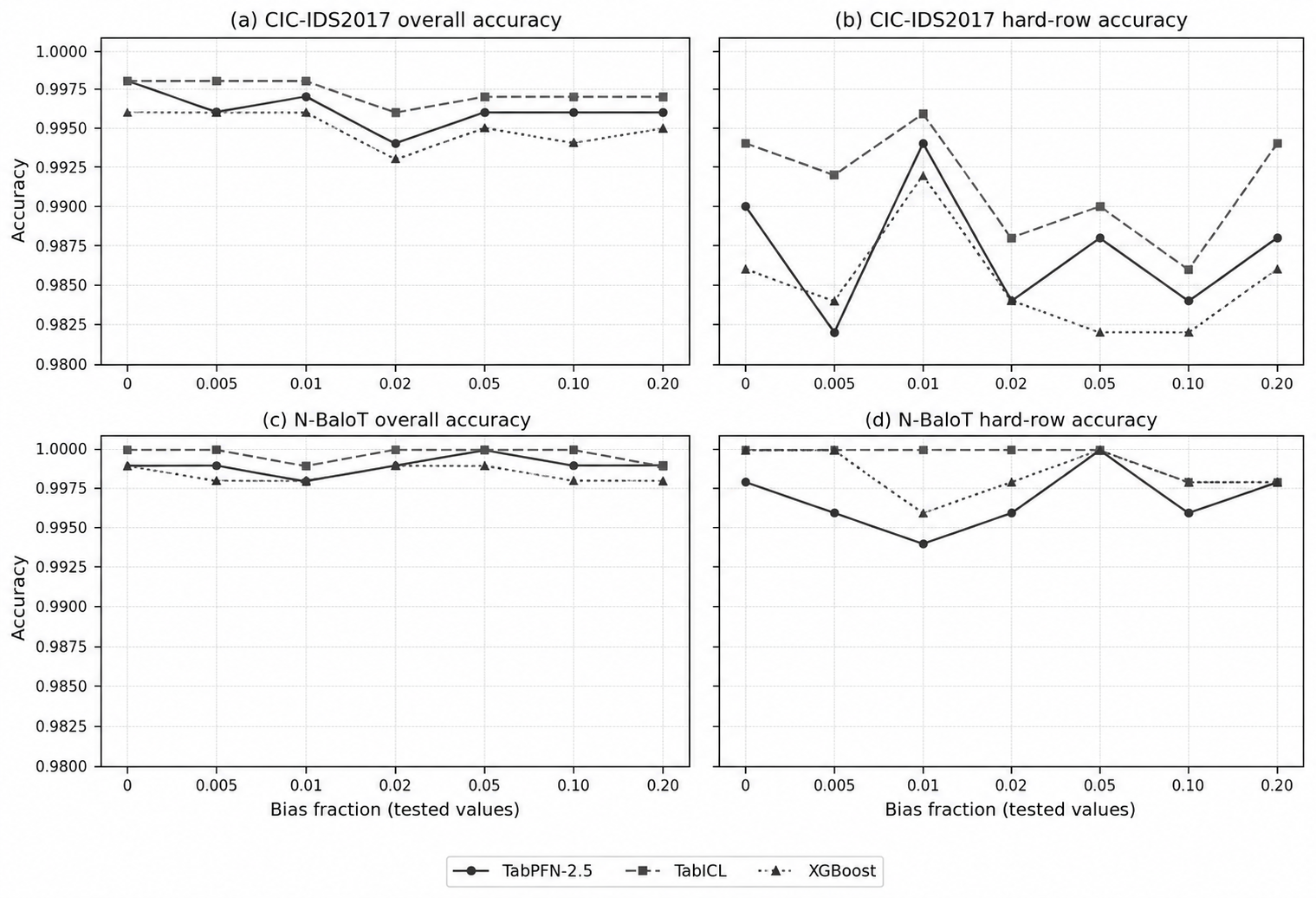}%
    }{%
        \fbox{\parbox[c][2.8cm][c]{0.80\textwidth}{\centering
        Missing figure: \texttt{Figures/context\_selection\_results.png}\\
        Rename the context-selection image to this filename.}}%
    }
    \caption{Overall and hard-row accuracy under boundary-crowding context
    selection on CIC-IDS2017 (top) and N-BaIoT (bottom). Points show means
    over five trials.}
    \label{fig:context-selection}
\end{figure}

Figure~\ref{fig:context-selection} reports overall and hard-row accuracy across
the full bias-fraction sweep.

On CIC-IDS2017, TabPFN-2.5 clean accuracy ranges from 0.994 to 0.998,
TabICL from 0.996 to 0.998, and XGBoost from 0.993 to 0.996. Hard-row
accuracy likewise shows no monotonic decline. On N-BaIoT, overall accuracy remains between 0.998 and 1.000 and
hard-row accuracy between 0.994 and 1.000 across models and bias fractions.

Thus, increasing the share of genuine rows nearest to the opposite class does
not produce a systematic degradation, even at 20\% bias and even when
analysis is restricted to the boundary-near hard-row subset. Under the tested
preprocessing and nearest-opposite-class boundary-crowding strategy, this
selection-only attack is therefore much weaker than direct context-integrity
violations such as incorrect labels or a shared trigger--target association.
This null result is specific to the evaluated boundary-crowding mechanism and
should not be interpreted as evidence that clean-label context selection is
ineffective in general.

\subsection{Evasion Results}

\begin{figure}[t]
    \centering
    \IfFileExists{Figures/evasion_results_final.png}{%
        \includegraphics[width=\textwidth]{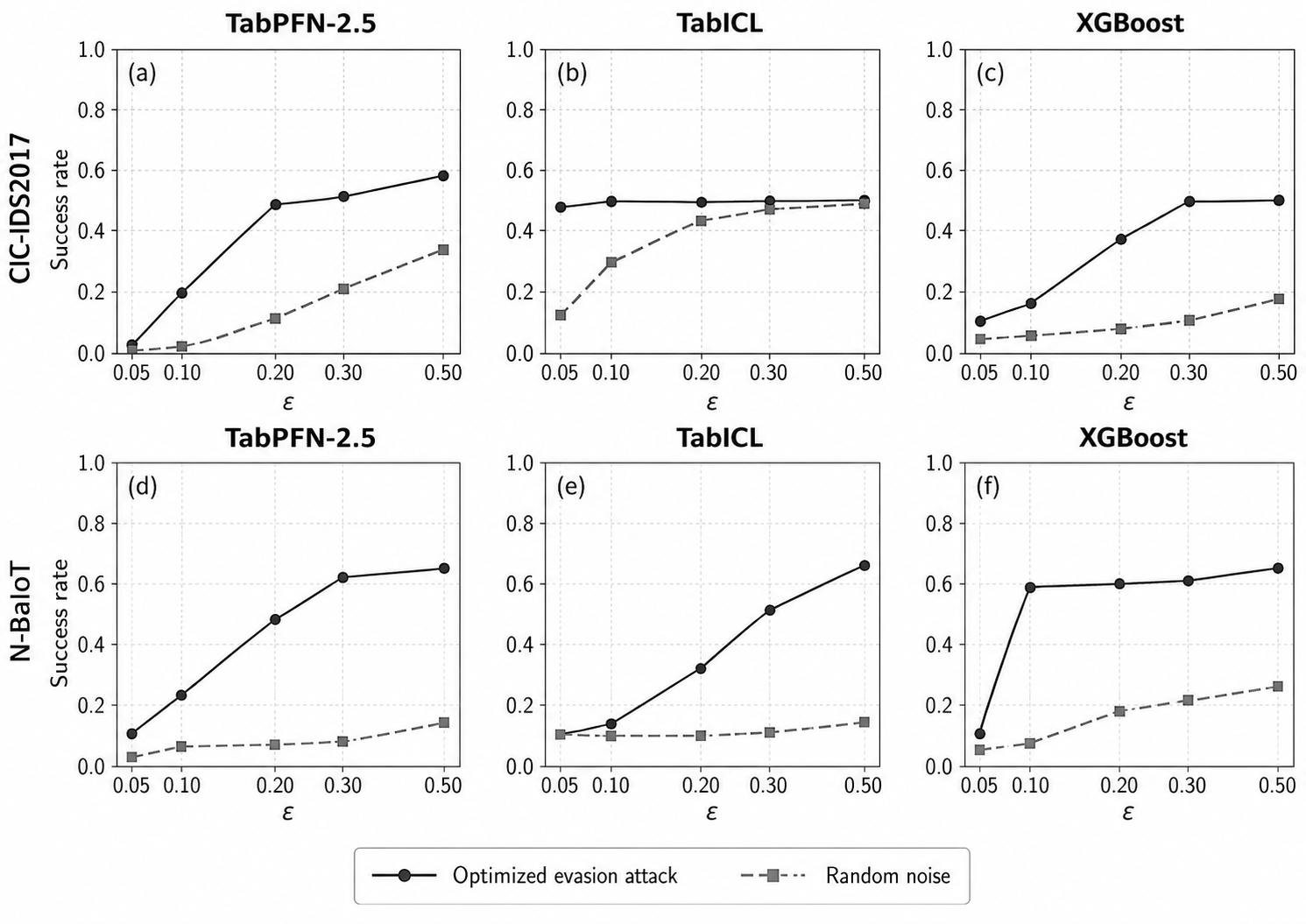}%
    }{%
        \fbox{\parbox[c][3.0cm][c]{0.84\textwidth}{\centering
        Missing figure: \texttt{Figures/evasion\_results\_final.png}\\
        Regenerate it from the $\epsilon=0.05$--$0.50$ results.}}%
    }
    \caption{Victim evasion rate and equal-budget random-noise rate across
    $L_\infty$ budgets for CIC-IDS2017 (top) and N-BaIoT (bottom). Columns
    show TabPFN-2.5, TabICL, and XGBoost. Points are five-trial means.}
    \label{fig:evasion-results}
\end{figure}

Figure~\ref{fig:evasion-results} compares the optimized evasion attack with
equal-budget random noise. Because all retained features are mutable in the
standardized feature space, these results quantify model-space susceptibility;
they do not show that every successful perturbation corresponds to realizable
network traffic.

Victim self-consistency is 1.000 in every evaluated setting, and surrogate
fidelity is approximately 0.98--0.99. These checks reduce the likelihood that
the observed effects are caused by nondeterministic victim predictions or a
poorly matched surrogate.

On CIC-IDS2017, TabPFN-2.5 reaches its largest attack--noise gap of 0.380 at
$\epsilon=0.20$, while XGBoost reaches 0.385 at $\epsilon=0.30$. TabICL shows
a different pattern: its gap is 0.351 at $\epsilon=0.05$ but falls to 0.006 at
$\epsilon=0.50$. At the larger budgets, random noise alone fools TabICL at
nearly the same rate as the optimized perturbation, so the high raw evasion
rate there is better interpreted as general perturbation sensitivity than as a
strong directional advantage.

On N-BaIoT, the optimized attack remains more clearly separated from random
noise. TabPFN-2.5 reaches a maximum gap of 0.543 at $\epsilon=0.30$.
TabICL's gap grows from 0.001 at $\epsilon=0.05$ to 0.514 at
$\epsilon=0.50$. XGBoost reaches its largest gap of 0.520 at
$\epsilon=0.10$ and remains above the random-noise control at every tested
budget. At $\epsilon=0.50$, the XGBoost evasion and random-noise rates
are 0.647 and 0.261, respectively, giving a positive gap of 0.386.

The XGBoost result clarifies the scope of this attack family. Unlike context
poisoning, model-space test-time evasion is not specific to in-context
learning: it affects the decision boundary of both in-context and conventional
tabular models. Practical network-level exploitability remains a separate
question because the present perturbations do not enforce traffic-generation
constraints.

\subsection{Cross-Attack Comparison}

Table~\ref{tab:cross-attack} summarizes the main experimental outcomes using
quantitative results where they are directly comparable.

\begin{table}[H]
\centering
\caption{Cross-attack summary.}
\label{tab:cross-attack}
\footnotesize
\renewcommand{\arraystretch}{1.15}
\setlength{\tabcolsep}{3.5pt}
\begin{tabularx}{\textwidth}{@{}lYYY@{}}
\toprule
\textbf{Attack} & \textbf{CIC-IDS2017} & \textbf{N-BaIoT} &
\textbf{Main interpretation} \\
\midrule
Label flipping & At $p=0.20$: $0.989\pm0.003$ (TabPFN-2.5), $0.882\pm0.073$ (TabICL) clean accuracy &
At $p=0.20$: $0.996\pm0.001$ (TabPFN-2.5), $0.869\pm0.061$ (TabICL) clean accuracy &
Wrong labels can degrade accuracy at large budgets; this experiment measures
broad degradation rather than trigger-conditioned targeted control. \\
\addlinespace[2pt]
Backdoor poisoning & At $p=0.005$: ASR $0.9592\pm0.0391$ (TabPFN-2.5), $0.9972\pm0.0030$ (TabICL) &
At $p=0.005$: ASR $1.0000\pm0.0000$ (TabPFN-2.5), $0.9972\pm0.0011$ (TabICL) &
Strong effect on in-context models; the clean-trained XGBoost negative
control is unaffected. \\
\addlinespace[2pt]
Context selection & At $b=0\rightarrow0.20$: $0.9976\pm0.0021\rightarrow0.9960\pm0.0031$ (TabPFN-2.5), $0.9978\pm0.0018\rightarrow0.9970\pm0.0028$ (TabICL), $0.9956\pm0.0013\rightarrow0.9954\pm0.0019$ (XGBoost) &
At $b=0\rightarrow0.20$: $0.9990\pm0.0007\rightarrow0.9988\pm0.0011$ (TabPFN-2.5), $0.9996\pm0.0005\rightarrow0.9992\pm0.0008$ (TabICL), $0.9986\pm0.0011\rightarrow0.9982\pm0.0020$ (XGBoost) &
Real, correctly labeled boundary-crowding rows do not provide comparable control. \\
\addlinespace[2pt]
Evasion & Largest mean gap across five trials: 0.385 (XGBoost, $\epsilon=0.30$) &
Largest mean gap across five trials: 0.543 (TabPFN-2.5, $\epsilon=0.30$) &
Optimized test-time perturbations affect both in-context and conventional models. \\
\bottomrule
\end{tabularx}
\end{table}

The cross-attack comparison shows that direct integrity violations provide
stronger control than the tested selection-only mechanism, while evasion must
be interpreted against equal-budget noise because high raw success can partly
reflect general perturbation fragility.

\subsection{Practical Implications}

Inference-time context should be treated as security-sensitive data, with
attention to provenance and label quality. For test-time robustness,
deployment claims should separate optimized attack effects from ordinary noise
sensitivity and ultimately enforce domain-valid feature constraints.

\subsection{Limitations and Future Work}

The evasion method uses surrogate transfer rather than true white-box
victim gradients and is a simplified constrained PGD variant inspired by
CAPGD, not the published CAPGD/CAA implementation. All retained features are
currently treated as attacker-controllable, and the largest standardized
budgets, especially $\epsilon=0.50$, may be unrealistic for some network
features. Future evaluations should enforce feature mutability, type, range,
and cross-feature constraints and compare the current attack with the full
CAPGD/CAA implementation under the same experimental setting.

The context-selection study evaluates only one realistic strategy---nearest-
opposite-class boundary crowding. Its null result therefore should not be
generalized to context-selection attacks as a whole. Class-imbalance,
temporal, device-specific, or diversity-based selection mechanisms may behave
differently and should be evaluated in future work. It would also be useful to test stronger adversaries
that can influence context composition while still being restricted to genuine,
correctly labeled rows.

The experiments use two network intrusion datasets on which all three models
achieve very high clean accuracy, so broader claims require additional, more
difficult, and more diverse tabular tasks. Future work should
therefore extend the evaluation to other security and non-security tabular
datasets and investigate whether the same differences between context-time and
test-time attacks remain. Additional work could also evaluate practical
mitigation strategies, such as context validation, provenance checks, and
constraint-aware preprocessing, without shifting the main focus away from the
attack analysis.

All experiments use public research datasets and do not target live networks
or organizations.
\section{Conclusion}
\label{sec:conclusion}

This paper compares four attacks on tabular models within one taxonomy based
on attack stage, label consistency, and distributional realism. Label flipping
mainly causes broad accuracy degradation, whereas backdoor context poisoning
can strongly redirect in-context predictions with little loss in clean
accuracy. The evaluated nearest-opposite-class boundary-crowding strategy,
which uses only genuine and correctly labeled rows, causes no systematic
degradation on CIC-IDS2017 or N-BaIoT up to a 20\% bias fraction, including
on the boundary-near hard-row subset; this finding is specific to that
selection strategy.

Test-time evasion affects TabPFN-2.5, TabICL, and XGBoost, but the random-noise
control changes how some results should be interpreted. In particular,
TabICL's optimized advantage on CIC-IDS2017 largely disappears at high
budgets, whereas the separation grows on N-BaIoT. Overall, the findings support
evaluating context-specific and decision-boundary attacks separately and
reporting attack success together with data realism and appropriate controls.
Future work should enforce realistic network-feature constraints, test
additional context-selection strategies, and extend the comparison to harder
and more diverse tabular datasets.

\subsubsection*{Disclosure of Interests.}
The authors have no competing interests to declare that are relevant to the content of this article.

\bibliographystyle{splncs04}
\bibliography{references}

\end{document}